\documentclass[runningheads]{llncs}
\usepackage{booktabs}
\usepackage{url}
\usepackage{graphicx}
\usepackage{csvsimple}
\usepackage{subcaption}
\begin{document}
\title{Enhanced Object Detection in Floor-plan through Super Resolution}
\titlerunning{Effects of Super-Resolution on Floor plan Object Detection}

\author{Dev Khare\inst{1}
N S Kamal\inst{2},
H B Barathi Ganesh \inst{1,2},
V Sowmya\inst{1}*,\\
V V Sajith Variyar\inst{1},}
\authorrunning{Dev et al.}
%
\institute{Center for Computational Engineering and Networking (CEN),\\ Amrita School of Engineering, Coimbatore,\\Amrita Vishwa Vidyapeetham, India\\
\email{dev.khare108@gmail.com,v\_sowmya@cb.amrita.edu,\\ vv\_sajithvariyar@cb.amrita.edu
}\\
\and
RBG.AI, Resilience Business Grids LLP, SREC Incubation Center, Coimbatore, Tamil Nadu, India\\
\email{kamal@rbg.ai,aiss@rbg.ai}\\
}
\maketitle              
\begin{abstract}
Building Information Modelling (BIM) software use scalable vector formats to enable flexible designing of floor plans in the industry. Floor plans in the architectural domain can come from many sources that may or may not be in scalable vector format. The conversion of floor plan images to fully annotated vector images is a process that can now be realized by computer vision. Novel datasets in this field have been used to train Convolutional Neural Network (CNN) architectures for object detection. Image enhancement through Super-Resolution (SR) is also an established CNN based network in computer vision that is used for converting low resolution images to high resolution ones. This work focuses on creating a multi-component module that stacks a SR model on a floor plan object detection model. The proposed stacked model shows greater performance than the corresponding vanilla object detection model. For the best case, the the inclusion of  SR showed an improvement of 39.47\%  in object detection over the vanilla network. Data and code are made publicly available at \UrlFont{https://github.com/rbg-research/Floor-Plan-Detection}.

\end{abstract}

\section{Introduction}
Design thinking in Architecture, Engineering and Construction (AEC) is adapting to new methods in Artificial Intelligence (AI) at a rapid pace. The process optimization capabilities of AI are fueling the need for automation of complex processes in the construction industry. There are many reasons for this: the need for critical testing of inferences before the project moves into stages where designs become concrete, a more general approach for influencing floor plan designs, and building architecture based on primary data; not biased by employee experience. A centralized database of relevant information comprising of regional interests and metrics would help formulate better inferences for design selection than just relying on individual expertise. However, the main challenge in this process lies in the data representation from sources like Computer Aided Design (CAD) tools, Building Information Modeling (BIM) software, etc \cite{b3}. For sketch-type data, data representation becomes very cumbersome as the information contains vector objects and geometric entities; for floor plan data, this is a big hurdle recently realized by Kalevro et al. resulting in the creation of the CubiCasa5k corpus \cite{b2}. CubiCasa5k is a corpus containing five thousand floor plan images and labels that segment floor plan images to find the polygons representing rooms, walls, and various icons in a floor plan like doors and windows. This process uses Convolutional Neural Networks (CNNs), which influenced most of the research works in computer vision.

Segmentation of floor plans needs to be precise for an end-to-end application. After observing the segmentation results of the empirical model for floor plans of various sizes, we concluded that there is a need for the CubiCasa approach to incorporate image enhancement as an essential preliminary step. Before digitisation, architects used to scan building plans manually resulting in different image qualities for different projects. Now, architectural drawings come in the form of vector images where everything is in a scalable format. Since a large chunk of floor plan images that are retrieved from web data have varying sizes, a workable data pipeline for floor plan images requires the use of image enhancement before storing the data into a database. Super-resolution is one such method used for image enhancement and is approached differently for various use cases. The process mainly varies with the type of image; for example, enhancing a facial appearance would require a more comprehensive approach with millions of variables and repeated steps. To address the question of end-to-end automation of floor plan image data to 3D building models, we explore this idea with experimental image enhancement results and its influence on floor plan object detection.

\section{Literature Review}

Automating floor plan image annotation was first looked at by Liu et al \cite{b1}. The approach proposed in the paper uses a network inspired by the ResNet-152 architecture. The problem is subdivided into three sections that are solved simultaneously; this makes the neural network a multi-tasking CNN. The multitasking aspect becomes clear after looking at the innovative use of junction point orientations in the form of heatmaps. This means of point detection allows for precise detection of the wall skeleton in floor plan images. The main drawback of this implementation was lack of data used for training, this was solved by the creation of the CubiCasa5k dataset. The neural architecture used by Kalevro et al. \cite{b2} contains multiple resnet blocks, similar to \cite{b1}. Their network uses a loss function containing room losses, icon losses, and a specially defined heatmap loss (containing junction point data) resulting in a multi-component output \cite{b2}. The future of building design estimation lies in the ability of models to manipulate core conceptual ideas in the design, like the size of rooms based on estimates of the number of people or the type of walls based on the region's cultural influences,logistical constraints,etc. The first step to achieving such feats is creating a successful base model for parametrizing floor plans, 3D point clouds for buildings, etc.\cite{b10}.

A major use case for floor plan annotation is that of 3D building reconstruction. In 2019, an overview of the state-of-the-art methods for converting 2D images to 3D was given by Han et al. \cite{b4}. The paper brought to notice a lot of exciting approaches for solving the 3D reconstruction problem. One such method was combining loss functions that measure different types of loss metrics in the model; this is a powerful approach commonly used for 3D reconstruction.\cite{b4}. 3D reconstruction is highly dependent on the available data. Reconstruction of 3D point clouds of buildings with complex designs, carried out by Zhang et al. \cite{b5}, is a recent achievement in this field that uses 2D building section data for its inference. Super-resolution in this implementation takes care of data loss when the sections are passed through the proposed network, enhancing the overall prediction accuracy. Training the proposed network on CAD models of building reconstructions meant that simple slicing operations were sufficient for section data creation. Another approach recently explored by Yu et al.\cite{b6} used topological heatmaps to reconstruct buildings directly from satellite images. This implementation used the WHU corpus to extract topological heatmaps as labels for training \cite{b6,b7,b8,b9}. A lot of these works rely on the output data being in a point cloud format which contains no information about the actual shape of the building. Floor plan wall annotation, in \cite{b1}, retrieves polygonal information that is editable in nature; using these polygons for 3D reconstruction takes a simple extrusion, making this approach more feasible \cite{b1}. Another advantage of using floor plan data is the volume of data on the web that can be used to fine tune such models. This brings us to the question of image quality.

The use of image enhancement as a preliminary step to classification or segmentation problems is applicable for many computer vision use cases\cite{b11}. Super-resolution networks are CNN based networks that are trained on high resolution images to serve the purpose of noise free image interpolation\cite{b12,b13}. Image enhancement is essential for precise segmentation of floor plans; this was made very clear in the application by Han et al. The use of a super-resolution framework called SRCNN enhances sectional 2D data for more precision on target output in their work \cite{b5}. There are multiple frameworks for super-resolution that exist. In 2017, Enhanced deep super-resolution network (EDSR) was the state of the art approach to super-resolution; the method used a standard ResNet based architecture and improved on several problems in model structure \cite{b15}. In 2021, the latest methods involve generative adversarial networks (GAN's) and probabilistic graphical models. Saharia et al. have improved the existing techniques with their iterative refinement method for image super-resolution\cite{b14}. By incorporating a Markov chain to add a parametrized gaussian noize to a low-resolution image (upsampled by bicubic interpolation), a posterior probability is defined to iteratively denoise gaussian noise resulting in a high-resolution image. In this work, the denoising step uses a modified UNet architecture with a loss function that is made aware of this gaussian diffusion process. The model surpasses all prior state-of-the-art super-resolution methods; however, the process is highly time-consuming for larger images. Therefore, it is not recommended to use the method in \cite{b14} for floor plan use cases as the desired high resolution image is quite large.

For a use case involving floor plan images, there are faster networks that perform super resolution with minimal drop in quality. In 2016, Shi et al \cite{b16} have used an efficient sub-pixel layer in their architecture for real time single image video enhancement. Similarily, Dong et al \cite{b17} worked on a faster version of the existing SRCNN by removing the process of bi-cubic interpolation before model evaluation; their network used an hourglass CNN instead of direct up-scaling to achieve this. Another notable work, by Lai et al \cite{b18}, on a laplacian pyramid based CNN architecture (LapSRN) outperforms both \cite{b16} and \cite{b17} in performance. The use of local skip connections and shared network parameters for different up-scaling factors makes this one of the most versatile networks for super-resolution.
 
\section{Methodology}

The application of deep learning in object detection can be challenging given the time constraints and system requirements. For the task of floor plan annotation, we have assumed that this process need not be real time. Since we are performing an end to end conversion, the main objective here is to enhance accuracy without constraints on time and computing power. To do handle this we experiment a novel multi-component module that performs image enhancement followed by the object detection. As mentioned earlier, CubiCasa5k is a corpus that contains five thousand annotated floor plans for the use of training a multitask CNN. From this corpus we select ninety low resolution floor plans and observe the increase in performance after super-resolution. Since CubiCasa5K is a robust corpus with varying sizes, we can hope for some improvement in performance of low resolution floor plans.

\subsection{Preliminary Step}
This work performs super-resolution for image enhancement before detecting floor plan icons and room types; stacking super-resolution frameworks with the CubiCasa architecture results in a multi-component module that does just this. The networks chosen for super-resolution here are EDSR, ESPCN, FSRCNN, and LapSRN \cite{b15,b16,b17,b18}.
For the inference, we need a quantifiable measure that could help us make a conclusive statement on improving performance by using super-resolution as a preliminary step. For this, the accuracy scores have to be based on ground truth scaled by the same factor chosen for super-resolution; the CubiCasa5k corpus provides ground truth in scalable vector graphics (SVG) format making this possible. 

\subsection{Wall/Room Detection}
\begin{figure}[htbp]
	\centering
	\includegraphics[trim={0 0.6cm 0 3cm},clip,width=0.7\textwidth]{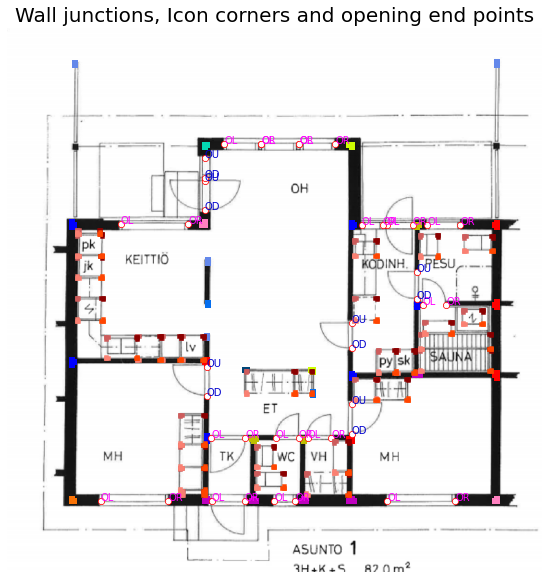}
	\caption{Junction points colour coded based on mapping to different labels}
	\label{fig:Membrane}
\end{figure}

Annotations are of three types here; rooms, icons, and junction points. Junction point information results in a heatmap that contains a mapping of junction point orientations for walls, icons, windows and doors. The predicted heatmaps are thresholded to retrieve the exact junction point locations. For this application, loss functions (defined by \cite{b1}) are used to optimize three segmentation problems. Icon, room, and heatmap losses were set up to improve the accuracy of a common goal: room and icon polygon detection.
The class labels for all three tasks were manually annotated. The junction points are colour coded based on their connections to rooms and icons. The CubiCasa5k dataset has ground truth information that comes in SVG format. It contains an archive of polygon types of which 12 are rooms (including walls) and 11 are icons (including doors and windows). A mapping is made to seperate 21 junction points into their respective heatmaps; it contains 13 wall junction orientations, 4 icon corner orientations and 4 window/door orientations.

\subsection{Post-processing}

The multi task CNN used for object detection in this paper uses many post processing techniques. The junction points retrieved from respective heatmaps get connected based on geometric orientations. Specifically for wall detection, the process results in a wall skeleton. The algorithm considers junction point triplets for definition. After this, a prunning step, defined to consider self intersecting polygons and incorrectly classified regions, makes the polygon detection more precise by using the segmentation maps as means for refinement. In Fig.2, for the room values alone we show the how the room detection looks with and without post processing. 

\begin{figure}[htbp]
	\centering
	\includegraphics[trim={0 0 0 0},clip,width=\textwidth]{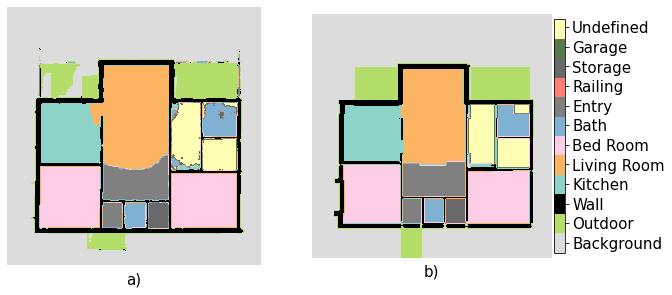}
	\caption{a) Segmented rooms b) Room polygons after post processing}
	\label{fig:Membrane}
\end{figure}

\section{Results and Discussion}
The model architecture in \cite{b2} is stacked with various SR models (refer section 3.1) to provide further insight into the functioning of this model; a comparative study is done on the SR inference latencies and performance metrics of the SR stacked models.

\subsection{Inference Latency}

EDSR, ESPCN, FSRCNN, and LapSRN \cite{b14,b15,b16,b17} can upscale any image by a factor of 2,3 or 4. However, with the upscaling factor comes an exponential rise in computational time as the number of network parameters are greater for large scaling factors. The surge in computational time can also come from large input image sizes, so we have limited the use of super-resolution to images of size less than 800x800 pixels for this application. Despite these modifications, on observation we find that EDSR is much slower than the other methods used in this paper. This is because EDSR contains 43 million learnable parameters which is considerably higher compared to ESPCN, FSRCNN and LapSRN. This is shown in table.1 .

\begin{table}
    \centering
    \caption{Execution time (in seconds) of SR networks chosen}
    \begin{tabular}{lrrrr}
\toprule
Image size&     EDSR &  ESPCN &  LapSRN &  FSRCNN \\
\midrule
690x769   &  215.431 &  0.340 &   2.245 &   0.320 \\
896x890   &  323.201 &  0.509 &   4.136 &   0.490 \\
1130x1016 &  353.921 &  0.797 &   4.907 &   0.686 \\
1449x1373 &  745.119 &  1.357 &   8.501 &   1.253 \\
\bottomrule
\end{tabular}
\end{table}
For the inference step, we found it fit to do a performance test on the influence of image enhancement(pre-processing) on object detection. The experiment was set up by stacking the super-resolution frameworks in series and then performing a scaling operation on the ground truth annotations for quantifying the results. For testing this process we have used 90 images from the CubiCasa5k corpus selected at random. We have chosen micro average as the performance metric in measuring the overall accuracy of the segmentation. Micro average precision, recall and F1 scores are helpful for testing the performance with data containing imbalanced class objects. Experiments were run on an AWS server with
NVIDIA T4 Tensor Core GPU having 320 Turing Tensor cores, 2,560 CUDA cores, and 16 GB of memory. Intel(R) Xeon(R) Platinum 8259CL CPU with a clock speed of 2.50GHz, 32 GB RAM(DDR4) and 8 vCPUs. 

\subsection{Results}
The multi-component module performed room detection with higher accuracy than that of when using CubiCasa5k model alone. Since wall detection is mapped to room detection we can say that there is improvement in the junction point detection and post processing. But the same cannot be said about icon detection. There is a slight performance drop in icon detection after performing super-resolution. We can attribute this to the fact that the icon heatmaps contain lesser no of points for each floor plan. For icon detection, there is more dependence on the data used for pre-training.
\begin{figure}[htbp]
	\centering
	\includegraphics[trim={0 0 0 0},clip,width=\textwidth]{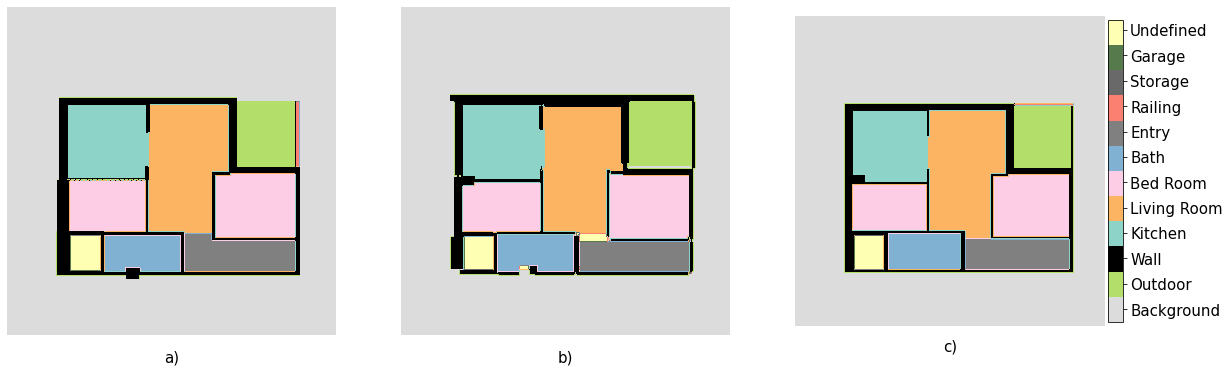}
	\caption{Comparision of a sample result with ground truth; a) Original image b) Pre-processed with EDSR c) Ground truth}
	\label{fig:Membrane}
\end{figure}
A major point to look at is that for images where icon junction point detection fails, post processing time raises by a huge amount. In these cases, there are a lot of icon points detected. The post processing algorithm removes self intersecting polygons and fine tunes the results based on segmentation maps. Since the time taken to remove self intersecting polygon is dependent on the number of icons detected, for failure cases the number of icons detected rise exponentially. After super-resolution the failure becomes even more catastrophic as there are even more polygons detected. Another point is that of polygon splitting. After super-resolution there are chances of polygons not joining properly to form the wall skeleton. This can be seen in Fig.3b . 

\begin{table}[htbp]
\centering
\caption{Room detection comparision of SR methods with original}
\resizebox{\textwidth}{!}
{\begin{tabular}{lccccccccccccccc}

\toprule
{} & \multicolumn{3}{l}{ESPCN} & \multicolumn{3}{l}{EDSR} & \multicolumn{3}{l}{FSRCNN} & \multicolumn{3}{l}{LapSRN} & \multicolumn{3}{l}{Original} \\
{} & precision & recall & f1-score & precision & recall & f1-score & precision & recall & f1-score & precision & recall & f1-score & precision & recall & f1-score \\
\midrule
Background  &     0.638 &  0.704 &    0.636 &     0.638 &  0.703 &    0.637 &     0.638 &  0.703 &    0.635 &     0.640 &  0.702 &    0.636 &     0.560 &  0.522 &    0.521 \\
Outdoor     &     0.532 &  0.537 &    0.459 &     0.542 &  0.537 &    0.458 &     0.542 &  0.536 &    0.469 &     0.554 &  0.535 &    0.469 &     0.282 &  0.593 &    0.227 \\
Wall        &     0.170 &  0.189 &    0.170 &     0.169 &  0.187 &    0.169 &     0.169 &  0.188 &    0.168 &     0.170 &  0.189 &    0.170 &     0.093 &  0.122 &    0.088 \\
Kitchen     &     0.336 &  0.482 &    0.304 &     0.336 &  0.483 &    0.304 &     0.338 &  0.483 &    0.304 &     0.346 &  0.485 &    0.309 &     0.153 &  0.650 &    0.142 \\
Living Room &     0.551 &  0.550 &    0.459 &     0.529 &  0.554 &    0.450 &     0.532 &  0.549 &    0.448 &     0.528 &  0.551 &    0.448 &     0.230 &  0.503 &    0.185 \\
Bed Room    &     0.481 &  0.558 &    0.449 &     0.470 &  0.557 &    0.437 &     0.481 &  0.557 &    0.448 &     0.467 &  0.555 &    0.435 &     0.287 &  0.708 &    0.238 \\
Bath        &     0.233 &  0.411 &    0.199 &     0.232 &  0.410 &    0.198 &     0.245 &  0.410 &    0.210 &     0.221 &  0.412 &    0.199 &     0.159 &  0.901 &    0.148 \\
Entry       &     0.270 &  0.457 &    0.253 &     0.269 &  0.456 &    0.251 &     0.268 &  0.453 &    0.251 &     0.278 &  0.460 &    0.255 &     0.207 &  0.924 &    0.207 \\
Railing     &     0.601 &  0.417 &    0.331 &     0.567 &  0.417 &    0.330 &     0.589 &  0.420 &    0.332 &     0.615 &  0.418 &    0.331 &     0.491 &  0.615 &    0.353 \\
Storage     &     0.494 &  0.577 &    0.438 &     0.494 &  0.577 &    0.438 &     0.493 &  0.577 &    0.437 &     0.495 &  0.576 &    0.438 &     0.479 &  0.887 &    0.455 \\
Garage      &     0.942 &  0.948 &    0.904 &     0.931 &  0.948 &    0.904 &     0.943 &  0.948 &    0.905 &     0.943 &  0.948 &    0.905 &     0.910 &  0.989 &    0.910 \\
Undefined   &     0.240 &  0.418 &    0.207 &     0.239 &  0.416 &    0.210 &     0.239 &  0.415 &    0.208 &     0.240 &  0.416 &    0.208 &     0.183 &  0.779 &    0.159 \\
micro avg   &     0.460 &  0.460 &    0.460 &     0.461 &  0.461 &    0.461 &     0.459 &  0.459 &    0.459 &     0.460 &  0.460 &    0.460 &     0.305 &  0.305 &    0.305 \\
\bottomrule
\end{tabular}}
\end{table}
\begin{table}[htbp]
\centering
\caption{Icon detection comparision of SR methods with original}
\resizebox{\columnwidth}{!}{
\begin{tabular}{lccccccccccccccc}
\toprule
{} & \multicolumn{3}{l}{ESPCN} & \multicolumn{3}{l}{EDSR} & \multicolumn{3}{l}{FSRCNN} & \multicolumn{3}{l}{LapSRN} & \multicolumn{3}{l}{Original} \\
{} & precision & recall & f1-score & precision & recall & f1-score & precision & recall & f1-score & precision & recall & f1-score & precision & recall & f1-score \\
\midrule
No Icon              &     0.939 &  0.931 &    0.935 &     0.939 &  0.931 &    0.935 &     0.939 &  0.931 &    0.935 &     0.939 &  0.931 &    0.935 &     0.950 &  0.933 &    0.941 \\
Window               &     0.155 &  0.115 &    0.109 &     0.132 &  0.115 &    0.097 &     0.134 &  0.116 &    0.099 &     0.151 &  0.113 &    0.106 &     0.095 &  0.115 &    0.025 \\
Door                 &     0.049 &  0.048 &    0.036 &     0.050 &  0.047 &    0.037 &     0.049 &  0.048 &    0.036 &     0.049 &  0.050 &    0.038 &     0.015 &  0.447 &    0.015 \\
Closet               &     0.214 &  0.226 &    0.159 &     0.212 &  0.228 &    0.159 &     0.213 &  0.227 &    0.158 &     0.212 &  0.224 &    0.157 &     0.106 &  0.617 &    0.094 \\
Electrical Applience &     0.112 &  0.196 &    0.119 &     0.114 &  0.204 &    0.122 &     0.113 &  0.199 &    0.120 &     0.115 &  0.202 &    0.122 &     0.082 &  0.626 &    0.070 \\
Toilet               &     0.168 &  0.228 &    0.117 &     0.157 &  0.227 &    0.106 &     0.174 &  0.220 &    0.099 &     0.169 &  0.228 &    0.117 &     0.182 &  0.898 &    0.159 \\
Sink                 &     0.134 &  0.161 &    0.081 &     0.111 &  0.161 &    0.079 &     0.130 &  0.160 &    0.081 &     0.135 &  0.160 &    0.081 &     0.159 &  0.761 &    0.159 \\
Sauna Bench          &     0.458 &  0.591 &    0.465 &     0.457 &  0.591 &    0.464 &     0.459 &  0.587 &    0.463 &     0.456 &  0.588 &    0.462 &     0.455 &  0.939 &    0.452 \\
Fire Place           &     0.911 &  0.866 &    0.843 &     0.911 &  0.866 &    0.843 &     0.911 &  0.867 &    0.844 &     0.911 &  0.867 &    0.844 &     0.920 &  0.966 &    0.909 \\
Bathtub              &     0.966 &  0.933 &    0.910 &     0.966 &  0.933 &    0.910 &     0.966 &  0.933 &    0.910 &     0.966 &  0.933 &    0.910 &     0.989 &  1.000 &    0.989 \\
Chimney              &     1.000 &  0.978 &    0.978 &     1.000 &  0.978 &    0.978 &     1.000 &  0.978 &    0.978 &     1.000 &  0.978 &    0.978 &     1.000 &  1.000 &    1.000 \\
micro avg            &     0.875 &  0.875 &    0.875 &     0.875 &  0.875 &    0.875 &     0.875 &  0.875 &    0.875 &     0.874 &  0.874 &    0.874 &     0.886 &  0.886 &    0.886 \\
\bottomrule
\end{tabular}}

\end{table}

\section{Conclusion and Future work}

The multi component module used in this paper was successful in improving the performance of icons and rooms in the CubiCasa5k framework. The best improvement in accuracy for the dataset chosen is 39.47\%; EDSR is the SR model used in this case. EDSR also showed the best result from all the super-resolution methods with a 12.17\% improvement on average. The improvement with LapSRN is  12.01\% on average but it is 82.76 times faster than EDSR, which makes up for the slight drop in performance. It is evident that LapSRN would be the best SR approach for practical use cases. On the context of time, for an end to end application there is some uncertainty. Existing post processing methods rely too much on the junction point search. This causes huge computation times for junction point failure cases where the search for self intersecting polygons becomes a combinatorially expensive process. There is definitely a need for better post processing methods to use the segment information more efficiently.

In the context of performance, we have used a scaling factor of 2 for all the methods presented here. There is a possibility that detection can improve even further with LapSRN and a higher scaling factor. However, it is clear that the stacked Super-Resolution method has to be used during training to alter the dimensions of low resolution images. This could enhance the overall performance of the network and push it to a wider range of use cases.

\end{document}